\documentclass[runningheads]{llncs}
\usepackage[T1]{fontenc}
\usepackage{microtype}
\usepackage[colorlinks,
            citecolor=gray,
            linkcolor=gray,
            urlcolor=gray,
            ]{hyperref}
\usepackage{orcidlink}

\usepackage{graphicx}
\usepackage{caption}
\usepackage{subcaption}
\usepackage{wrapfig}
\usepackage{url}
\usepackage{amsmath}
\usepackage{amssymb}
\usepackage{booktabs}
\usepackage{multirow}

\newcommand{\R}{\mathbb{R}}

\newcommand{\x}{\mathbf{x}}

\newcommand{\z}{\mathbf{z}}

\let\svthefootnote\thefootnote
\newcommand\freefootnote[1]{%
  \let\thefootnote\relax%
  \footnotetext{#1}%
  \let\thefootnote\svthefootnote%
}

\newcommand{\citep}[1]{\cite{#1}}
\newcommand{\citet}[1]{\cite{#1}}

\begin{document}
\title{Self-Supervised Siamese Autoencoders}
\author{
Friederike~Baier\inst{1} \and
Sebastian~Mair\inst{2} \orcidlink{0000-0003-2949-8781} \and
Samuel~G.~Fadel\inst{3} \orcidlink{0000-0002-4459-4336}
}
\authorrunning{Baier et al.}
\institute{Leuphana University of Lüneburg, Germany \and
Uppsala University, Sweden \and
Linköping University, Sweden \\ \email{samuel@nihil.ws}
}
\maketitle
\freefootnote{This paper was published in Advances in Intelligent Data Analysis XXII.}
\begin{abstract}
In contrast to fully-supervised models, self-supervised representation learning only needs a fraction of data to be labeled and often achieves the same or even higher downstream performance.
The goal is to pre-train deep neural networks on a self-supervised task, making them able to extract meaningful features from raw input data afterwards.
Previously, autoencoders and Siamese networks have been successfully employed as feature extractors for tasks such as image classification.
However, both have their individual shortcomings and benefits.
In this paper, we combine their complementary strengths by proposing a new method called SidAE (Siamese denoising autoencoder).
Using an image classification downstream task, we show that our model outperforms two self-supervised baselines across multiple data sets and scenarios.
Crucially, this includes conditions in which only a small amount of labeled data is available.
Empirically, the Siamese component has more impact, but the denoising autoencoder is nevertheless necessary to improve performance.
\end{abstract}

\section{Introduction}
Fully-supervised machine learning models usually require large amounts of labeled training data to achieve state-of-the-art performance.
For many domains, however, labeled training data is often costly and more challenging to acquire than unlabeled data.
Existing unsupervised or self-supervised pre-training approaches successfully reduce the amount of labeled training data needed for achieving the same or even higher performance \citep{jing2020self}.
Generally speaking, the aim is to pre-train deep neural networks on a self-supervised task such that after pre-training, they can extract meaningful features from raw data, which can then be used in a so-called \emph{downstream task} like classification or object detection.

In the context of image recognition, earlier work focuses on designing specific pretext tasks.
These include, e.g., generation-based methods such as
inpainting \citep{pathak2016context},
colorization \citep{zhang2016colorful},
and image generation (e.g., with generative adversarial networks \citep{goodfellow2014generative}),
context-based methods involving Jigsaw puzzles \citep{noroozi2016unsupervised},
predicting the rotation angle of an image \cite{gidaris2018unsupervised},
or relative position prediction on patch level~\citep{doersch2015unsupervised}.
Yet, resulting representations are rather specific to these pretext tasks, suggesting more general semantically meaningful representations should be invariant to certain transformations instead of covariant \citep{misra2020self,chen2020simple}.

Following this argument, many recent state-of-the-art models have been built on Siamese networks \cite{chen2020simple,he2020momentum,grill2020bootstrap,chen2021exploring}.
The idea is to make the model learn that two different versions of one entity belong to the same entity and that the factors making the versions differ from each other do not play a role in its identification.
The simple Siamese (SimSiam) model \citet{chen2021exploring} is designed to solve the common representation collapse issue arising from that strategy.

Another family of models used for pre-training neural networks, acting as feature extractors, are autoencoders.
They build on the principle of maximizing the mutual information between the input and the latent representation \citep{vincent2010stacked}.
However, good features should not contain as much information as possible about the input, but only the most relevant parts \citep{tian2020makes}.
This is commonly done by keeping the dimensionality of the latent space lower than that of the input and by adding noise to the image. 
This is the idea of a denoising autoencoder \cite{vincent2008extracting}.

By themselves, both approaches to learning representations lead to simple and effective solutions.
This simplicity allows us to leverage both approaches to design an arguably even more powerful feature extractor, as both strategies result in non-conflicting ways of learning representations.

In this paper, we propose the combination of a Siamese network and a denoising autoencoder to create a new model for self-supervised representation learning that comprises advantages of both and could therefore compensate for the shortcomings of the individual components.
We introduce a new model named SidAE (\underline{Si}amese \underline{d}enoising \underline{a}uto\underline{e}ncoder), which aims to adopt the powerful learning principles of both Siamese networks with multiple views of the sample input and denoising autoencoders with noise tolerance.
Our experiments show that SidAE outperforms its composing parts in downstream classification in a variety of scenarios, either where all labeled data is available for the downstream task or only a fraction of it, both with or without fine-tuning of the feature extractor.
\looseness=-1

\section{Self-Supervised Representation Learning}
\label{sec:Selfsupervised}

Self-supervised learning leverages vast amounts of unlabeled data.
The key is to define a loss function using information that is extracted from the input itself.
Generally, self-supervision can be used in various domains, including text, speech, and video.
In this paper, we focus on the domain of image recognition.

We consider two stages:
(i) self-supervised pre-training and
(ii) a supervised downstream task.
The goal of the pre-training stage is to train a so-called \emph{encoder} neural network which learns to extract meaningful information from raw input signals.
In the second stage, we utilize the encoder as a feature extractor.
Then, a classifier, e.g., a simple fully-connected layer, is trained on top of the representations extracted by the encoder using a supervised loss and labeled data.
The weights of the encoder can be fine-tuned (optimized) or kept frozen (not optimized) for the downstream task.

The main challenges are determining which features are meaningful and how to tweak models to produce representations that satisfy certain properties.
It has been suggested that representations should contain as much information about the input as possible (InfoMax principle; \cite{linsker1989generate}).
Yet, others argue that task-specific representations should only contain as much signal as needed for solving a specific task, as task-unrelated information could be considered noise and might even harm performance (InfoMin principle; \cite{tian2020makes}).
In general, representations should be invariant under certain transformations, i.e., so-called \emph{nuisance factors}, such as lighting, color grading, and orientation, as long as those factors do not change the identity of an object.
One line of reasoning is that a model that can predict masked parts or infer properties of the data should have some kind of \emph{higher-level understanding} of the content \cite{henaff2020data,assran2023self}.

\begin{figure}[t]
  \includegraphics[width=\textwidth]{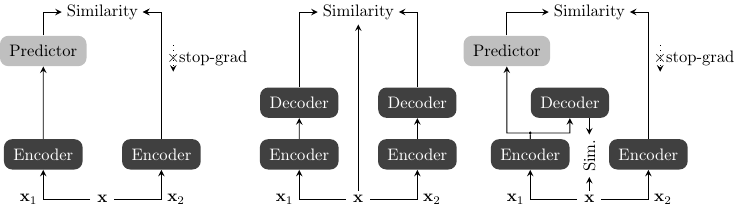}
  \caption{Siamese networks (left) and (denoising) autoencoders (middle), compared to our proposed model SidAE (right). The left and right illustrations should be symmetric, but we show only one side of the symmetry for brevity. The middle illustration shows two views due to our experimental setting (for a fair comparison), although a vanilla denoising autoencoder usually has only one.}
  \label{fig:models}
\end{figure}

\paragraph{Siamese networks.}
Siamese (also known as joint-embedding \cite{assran2023self}) architectures are designed in a way that a so-called \emph{backbone network} should learn that two distorted versions of an image still represent the same object.
Intuitively, the reasoning is that the features extracted from the same entity should be similar to each other, even under those distortions.

Our proposed model builds on SimSiam \cite{chen2021exploring}.
Due to its simplicity compared to other Siamese architectures, it is well-suited as a basis for more complex architectures.
In general, Siamese networks consist of two branches encoding two different \emph{views}, $\x_1$ and $\x_2$, of the same input $\x$, which we detail below.
Each branch in SimSiam contains both an encoder and a so-called \emph{predictor network}, where the encoders for the two branches share their weights.
In other approaches such as in MoCo \citep{he2020momentum} and BYOL \citep{grill2020bootstrap}, the weights of one network are a lagging moving average of the other.
Figure~\ref{fig:models} (left) illustrates the architecture of SimSiam, showing the stop-gradient operation only for one side for brevity.

In SimSiam and other state-of-the-art models, the encoder comprises a neural network, often referred to as \emph{backbone}.
In some cases, there is also an additional so-called \emph{projector network}.
A projector is also used in SimCLR \cite{chen2020simple}, where it improves the downstream performance.

The first step in pre-training is to extract two different \emph{views} of an (vectorized) input image $\x \in \R^{d_{\text{in}}}$.
These views are obtained by applying two sets of augmentations, $t_1$ and $t_2$, randomly sampled from a family of augmentations $T$.
The encoder $\operatorname{Enc}(\cdot)$ and predictor $\operatorname{Pred}(\cdot)$ networks are then used as
\begin{align*}
  \x_j = t_j(\x), \quad
  \z_j = \operatorname{Enc}(\x_j), \quad
  \mathbf{p}_j = \operatorname{Pred}(\z_j),
\end{align*}
where $\x \in \R^{d_{\text{in}}}, \x_j \in \R^{d_{\text{in}}}$, $\z_j \in \R^{d_{\text{hid}}}$, $\mathbf{p}_j \in \R^{d_{\text{hid}}}$.
Here, $d_{\text{in}}$ is the input dimensionality, $d_{\text{hid}}$ the hidden or latent dimensionality, and $j \in \{1, 2\}$.
Using $d_{\text{hid}} = \frac{1}{4} d_{\text{in}}$ was found to lead to better results when compared to $d_{\text{hid}} = d_{\text{in}}$ \cite{chen2021exploring}.

With this, the model can be trained by minimizing a distance $D(\mathbf{p}_i, \z_j)$ with $i, j \in \{1, 2\}$ and $i \neq j$.
Hence, the loss is minimized by forcing the representations to be close to each other.
Both negative cosine similarity and cross-entropy can be used as $D(\cdot, \cdot)$, but earlier reports show that the former leads to superior results \cite{chen2021exploring}.
Crucially, a stop-gradient operation is applied on $\z_j$.
Thus, while backpropagating through one of the branches, the other branch is treated as a fixed transformation.
More specifically, the loss for SimSiam is calculated as
\looseness=-1
\begin{align}
L_{\operatorname{si}} = \frac{1}{2} D(\mathbf{p}_1, \operatorname{stopgrad}(\z_2)) + \frac{1}{2} D(\mathbf{p}_2, \operatorname{stopgrad}(\z_1)),
\label{eq:Loss_Simsiam}
\end{align}
 where $\operatorname{stopgrad}(\cdot)$ denotes the stop-gradient operation.
This is essential to avoid collapsing solutions \cite{chen2021exploring} and works as a replacement for earlier attempts, such as using one branch with weights as a lagging moving average of the other \citep{he2020momentum,grill2020bootstrap}. 

\paragraph{Denoising autoencoders.}

Autoencoders are another family of self-supervised models and build on the concept of maximizing the mutual information between some input $\x$ and corresponding latent code $\z$ \citep{vincent2010stacked}.
Yet, they need to be restricted in some way such that the model does not simply learn an identity mapping where $\z$ is equal to $\x$, usually achieved by using a smaller dimensionality for $\z$.
For a denoising autoencoder, the input $\x$ is distorted, e.g., by applying Gaussian noise, and this corrupted version $\tilde{\x}$ is used as input to the model
$\x' = \operatorname{Dec}(\operatorname{Enc}(\tilde{\x}))$.
The loss $L_{\text{dae}} = D(\x', \x)$ is minimized to make $\x$ and $\x'$ as similar as possible.
Note that the denoising autoencoder does not see $\x$, only its corrupted version $\tilde{\x}$, but still aims to reconstruct $\x$.
Hence, the model should learn to focus more on the crucial information to reconstruct the original inputs, and optimally, the representations encode only relevant signals without noise.
\looseness=-1

\section{A Siamese Denoising Autoencoder}
\label{sec:Model}

The newly proposed model contains parts of SimSiam \citep{chen2021exploring} and a denoising autoencoder.
The design of the denoising autoencoder was inspired by works of \citet{li2017universal}.
To the best of our knowledge, the exact architecture and application as shown here have not yet been presented.
Note that components that appear in multiple models (i.e., the encoder, decoder, and predictor networks) are the same in SimSiam, SidAE, and the denoising autoencoder.

\subsection{Motivation}

Usually, a denoising autoencoder is trained on a single corrupted view of the data, while Siamese networks use two views.
Additionally, Siamese networks do not aim to reconstruct the original inputs from the information its encoder extracted, while the denoising autoencoder does.
We hypothesize combining both approaches results in a more powerful method since they compensate for the shortcomings of each other.
This difference in behavior can be used to the benefit of self-supervision, i.e., allowing the denoising autoencoder to access two different views of the same sample.
Additionally, the encoder used by the Siamese part is encouraged to keep enough information for the decoder to reconstruct the inputs.
This symbiosis is the fundamental design principle in SidAE.

Intuitively, in contrast to pre-training a network with SimSiam, using SidAE should make the network learn that two views come from the same object.
In SimSiam, the original input is not directly used to optimize the model, as only augmented versions are fed into the model, and the loss is computed using representations in the latent space.
In SidAE, however, disrupted local information is mapped to the original, undisrupted global information by the denoising autoencoder part.
Hence, the model is explicitly encouraged to encode information about the original input $\x$ into the latent representations $\z_1$ and $\z_2$.

With this, compared to a denoising autoencoder, SidAE should not just be optimized to encode as much information as possible from the original input into the latent codes (while being robust to noise).
It should also ignore specific nuisance factors, which is achieved by the Siamese part of its network.
This should prevent the features from containing irrelevant details.
In what follows, we will describe the architecture and specify the loss function of SidAE.
A visual overview of SidAE is given in Figure~\ref{fig:models} (right).

\subsection{Architecture}

\paragraph{Input.}
\label{sec:Augmentations}
SidAE uses two views, $\x_1$ and $\x_2$, which are obtained by applying two augmentations, $t_1$ and $t_2$, which we sample from a set of augmentations $T$.
We employ the same augmentation pipeline as in SimSiam \cite{chen2021exploring} for CIFAR-10 which consists of the following transformations stated in PyTorch \cite{paszke2019pytorch} objects:
(i) \texttt{ColorJitter} with $\operatorname{brightness}=0.4$, $\operatorname{contrast}=0.4$, $\operatorname{saturation}=0.4$, and $\operatorname{hue}=0.1$ applied with probability $p=0.8$,
(ii) \texttt{RandomResizedCrop} with scale in $[0.2,1]$,
(iii) \texttt{RandomGrayscale} applied with probability $p=0.2$, and
(iv) \texttt{RandomHorizontalFlip} applied with probability $p=0.5$.
For Fashion-MNIST and MNIST, we additionally apply a \texttt{GaussianBlur} with a standard deviation in $[0.1,2.0]$ with probability $p = 0.5$.

\begin{figure}[t]
  \includegraphics[height=1.75cm]{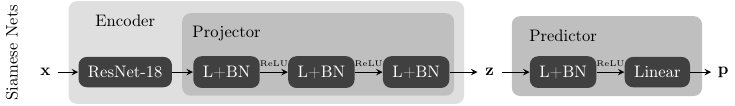}
  \includegraphics[height=1.75cm]{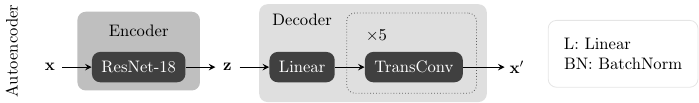}
  \caption{The components of the models used in our experiments.}
  \label{fig:components}
\end{figure}

\paragraph{Encoder.}
The backbone network in the encoder is a ResNet-18 \citep{he2016deep}.
As in SimSiam, we replace the last fully-connected layer with a 3-layer projection MLP.

\paragraph{Decoder.}
After the inputs $\x_1$ and $\x_2$ are encoded into the latent representations $\z_1$ and $\z_2$, they are fed into the decoder, producing the reconstructions $\x'_1$~and~$\x'_2$.
The decoder mirrors the encoder in that it up-samples the data step-by-step to reconstruct an output of the original input size. 
For this purpose, a fully-connected layer followed by five transposed convolution layers are stacked on each other.
Their kernels have a size of $3 \times 3$, [output] padding of 1, and stride of 2.
\looseness=-1

\paragraph{Predictor.}
The predictor is a two-layer MLP, the same as in SimSiam \citep{chen2021exploring}.
It maps the latent representations $\z_1$ and $\z_2$ to another latent space to produce two embeddings $\mathbf{p}_1$ and $\mathbf{p}_2$.
Both $\z_i$ and $\mathbf{p}_j$ ($i, j \in \{ 1, 2 \}$ and $i \neq j$) have the same dimensionality since they are compared to each other later.
We depict the full structure with the encoder, projector, decoder, and predictor in Figure~\ref{fig:components}.

\paragraph{Loss.}
The loss for SidAE comprises two terms, one for the Siamese and one for the autoencoder part of the network.
The former is applied for comparing $\z_1$ to $\mathbf{p}_2$ and $\z_2$ to $\mathbf{p}_1$, and the latter compares the reconstructions $\x_1'$ and $\x_2'$ to the raw inputs, $\x_1$ and $\x_2$, respectively.
In SimSiam, the subcomponents of the loss are scaled by $\frac{1}{2}$, which amounts to scaling the whole loss by $\frac{1}{2}$.
We do the same for the autoencoder part of the loss.
Additionally, we introduce a weighting term on the Siamese (si) and autoencoder (dae) parts of the loss by setting a parameter $w \in [0,1]$.
This parameter controls the magnitude of the relative importance of each mechanism in the model.
The full loss $L_{\operatorname{sidae}}$ of SidAE is given by
\looseness=-1
\begin{align*}
L_{\operatorname{dae}} &= \frac{1}{2} D_{\operatorname{mse}}(\x, \x'_1) + \frac{1}{2} D_{\operatorname{mse}}(\x, \x'_2), \\
L_{\operatorname{si}} &= \frac{1}{2} D_{\operatorname{ncs}}(\mathbf{p}_1,\operatorname{stopgrad}(\z_2)) + \frac{1}{2} D_{\operatorname{ncs}}(\mathbf{p}_2,\operatorname{stopgrad}(\z_1)), \\
L_{\operatorname{sidae}} &= w L_{\operatorname{dae}} + (1-w) L_{\operatorname{si}},
\end{align*}
where $D_{\operatorname{mse}}$ is a mean squared error and $D_{\operatorname{ncs}}$ is the negative cosine similarity.

\section{Experiments}

We now perform several experiments regarding our proposed SidAE model, compare it against relevant baselines, evaluate the influence of the parameter $w$, and assess how the amount of pre-training impacts a downstream classification task on several real-world data sets.
For clarity and to avoid repetitiveness, we refer below to a denoising autoencoder whenever an autoencoder is mentioned.

\subsection{Experimental Setup}

\paragraph{Data.}
We evaluate on four real-world data sets: \emph{CIFAR-10} \citep{krizhevsky2009learning}, \emph{MNIST} \citep{lecun1995learning}, \emph{Fashion-MNIST} \cite{xiao2017fashion}, and \emph{STL-10} \citep{coates2011analysis}.
All contain images from ten classes and come with pre-defined train/test splits.
STL-10 is designated for self-supervised or unsupervised pre-training.
Thus, we use the unlabeled data for pre-training our models and the train and test sets for training and evaluating the model on the classification task.
For the other data sets, we use the same training protocol as \cite{chen2021exploring}.
We utilize the training set for self-supervised pre-training and for training the classifier as the downstream task.
In addition, we simulate harder tasks in which we only use a small fraction (i.e., 1\%) of the labeled training data for supervised training on the downstream task.
The subsets for each data set are drawn uniformly at random but are kept the same throughout all runs to enable a fair comparison.
We always use the full training set (or the unlabeled set in the case of STL-10) for pre-training.
To use the same architectures for all data sets, we resize all images to a resolution of 32$\times$32.

\paragraph{Baselines.} We compare SidAE against \emph{SimSiam}~\cite{chen2021exploring} and a \emph{denoising autoencoder}~\cite{vincent2008extracting}.
The encoder of the autoencoder uses a ResNet-18 with a three-layer projection MLP as used in SimSiam and SidAE and the decoder is like in SidAE.
The latent dimensionality of SidAE is set to $d_{\text{hid}}=2048$ \cite{chen2021exploring}.
Since SidAE sees two noisy views at a time, we also provide the autoencoder with two noisy views that are constructed in the same way.
The autoencoder uses the MSE loss.
We also compare against a fully-supervised ResNet-18 for the classification task.

\paragraph{Setup.} We use Python and implement all models in PyTorch \citep{paszke2019pytorch}.
All experiments run on a machine with two 32-core AMD Epyc CPUs, 512 GB RAM, and an NVIDIA A100 GPU with 40GB memory.
We re-implement SimSiam based on~\cite{chen2021exploring} and use the same settings for pre-training.
As an optimizer, we use stochastic gradient descent using a cosine decay schedule with an initial learning rate of 0.03, momentum of 0.9, weight decay of 0.0005, and batch size of 512.

We pre-train the backbone using SidAE, SimSiam, and a denoising autoencoder for 200 epochs.
The frozen pre-trained backbones are used as feature extractors for our supervised downstream task of image classification.
Each pre-trained model is evaluated at different stages of pre-training, i.e., in a range from 25 to 200 epochs, to assess how the duration of pre-training affects the quality of learned representations.
We replace the projector MLP within the backbone for the classifier with a fully-connected layer with an input dimension equal to~$\z$.
The classifier is trained for 50 epochs using a stochastic gradient descent with a learning rate of 0.05, a momentum of 0.9, and a batch size of 256.
Since the backbone weights are frozen, only the weights of the last fully-connected layer are trained.
We later evaluate the scenario of fine-tuning the frozen weights.

\subsection{Results}

Figure~\ref{fig:cifar} shows the effect of pre-training in downstream classification on \mbox{CIFAR-10}.
The more pre-training epochs we use, the better the downstream accuracy.
In the case where we only have 1\% of labeled training data for the classification task, only 25 pre-training epochs of the self-supervised methods are enough to outperform a fully-supervised classifier.
When 100\% of the labeled training data is available, it does not necessarily outperform a supervised baseline without fine-tuning the encoder.
Notably, the performance gap for the supervised baseline between 100\% and 1\% is way higher than for the self-supervised models.
These observations support that supervised models need a large amount of training data, performing rather poorly if only little labeled data is available.
Note that SimSiam yields the largest standard errors, whereas SidAE is more stable.
Overall, SidAE (with $w=0.5$) outperforms both the autoencoder and SimSiam.

\begin{figure}[t]
    \centering
    \includegraphics[width=.425\textwidth]{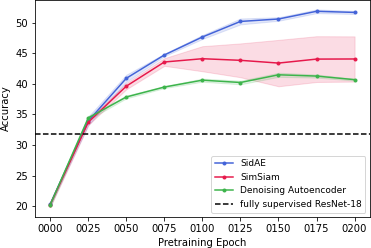}
    \hfill
    \includegraphics[width=.425\textwidth]{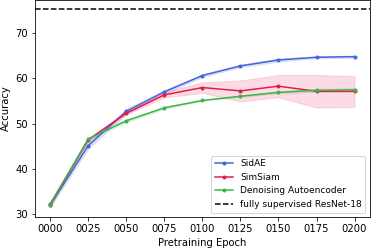}
    \caption{
    Classification accuracy on CIFAR-10 (averaged over 5 runs incl. std. err.) after different pre-training stages using a frozen pre-trained backbone. For downstream training, 1\% (left) and 100\% (right) of training data are used.
    }
    \label{fig:cifar}
\end{figure}

\begin{figure}[t]
  \begin{minipage}{.425\textwidth}
    \includegraphics[width=\textwidth]{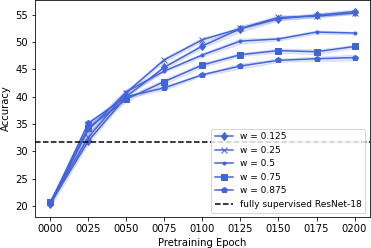}
    \caption{SidAE: The influence of the weight $w$ on CIFAR-10.
    }
    \label{fig:CIFAR-10_weight_w}
  \end{minipage}
  \hfill
  \begin{minipage}{.55\textwidth}
    \renewcommand{\figurename}{Table}
    \setcounter{figure}{0}
    \caption{Classification accuracies for SidAE \mbox{($w = 0.5$)}, SimSiam, and a denoising autoencoder averaged over five runs (std. err.).}
    \label{tab:modelcomp}
    \setcounter{table}{1}
    \setcounter{figure}{4}
    \renewcommand{\figurename}{Figure}
    \resizebox{\textwidth}{!}{
    \begin{tabular}[c]{cccccc}
        \toprule
          \textbf{Data}
        & \textbf{Pre-Train}
        & \multicolumn{1}{c}{\textbf{SidAE}}
        & \multicolumn{1}{c}{\textbf{SimSiam}}
        & \multicolumn{1}{c}{\textbf{Autoencoder}} \\
        \midrule

        {\multirow{2}{*}{CIFAR-10}}
         & 100\% & \textbf{64.75} (0.27) &  57.08 (3.16) & 57.44 (0.31) \\
         & 1\% & \textbf{51.67} (0.28)  &  44.06 (3.74) & 40.67 (0.25) \\
         \cmidrule(l){1-5}

        {\multirow{2}{*}{Fashion-MNIST}}
         & 100\% & \textbf{88.78} (0.12) &  85.10 (0.09) & 87.67 (0.05) \\
         & 1\% & \textbf{81.62} (0.22)  &  77.84 (0.19) &  80.20 (0.31) \\
         \cmidrule(l){1-5}

        {\multirow{2}{*}{MNIST}}
         & 100\% & \textbf{97.80} (0.13) &  96.59 (0.60) & 97.23 (0.11) \\
         & 1\% & \textbf{94.65} (0.49)  &  79.42 (5.15) &  88.41 (0.46) \\
         \cmidrule(l){1-5}

        {\multirow{1}{*}{STL-10}}
         & STL-10 & \textbf{61.18} (0.22) & 56.74 (0.74) &  48.39 (0.22) \\
        & CIFAR-10 & \textbf{66.20} (0.31) &  56.64 (0.25) & 57.08 (3.16) \\

        \bottomrule
    \end{tabular}
    }
  \end{minipage}
\end{figure}

\begin{figure}[t]
    \centering
    \includegraphics[width=.425\textwidth]{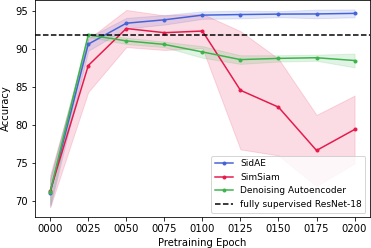}
    \hfill
    \includegraphics[width=.425\textwidth]{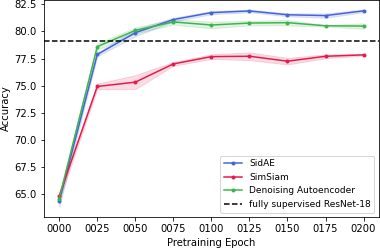}
    \caption{Classification accuracies on MNIST (left) and Fashion-MNIST (right) using 1\% of the training data for downstream training.}
    \label{fig:Comp_mnist_fmist}
\end{figure}

\begin{figure}[t]
    \centering
    \includegraphics[width=.425\textwidth]{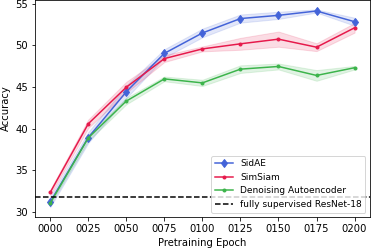}
    \hfill
    \includegraphics[width=.425\textwidth]{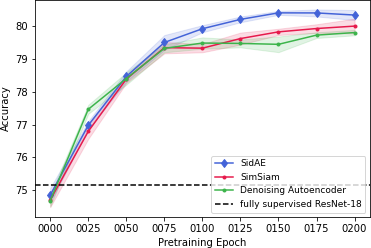}
    \caption{
      Classification accuracies on the fine-tuned downstream task (CIFAR-10) at different pre-training stages using encoders pre-trained with different models using 1\% (left) or 100\% (right) of the training data for the downstream task.
    }
    \label{fig:Finetuning}
\end{figure}

\paragraph{The Influence of $w$ in SidAE.}
We now investigate the influence of the weight~$w$ within the loss.
Figure~\ref{fig:CIFAR-10_weight_w} depicts the influence for the 1\% case on CIFAR-10.
The performance of the classifier when using SidAE with $w\in\{0.125,0.25\}$ is indeed better than the previously used choice of $w=0.5$.
The results are similar for the 100\% case and hence not shown.

\paragraph{Other data sets.}
Table~\ref{tab:modelcomp} shows the results after 200 pre-training epochs for the other data sets.
As before, SidAE outperforms its baselines.
Figure~\ref{fig:Comp_mnist_fmist} depicts the results for MNIST and Fashion-MNIST using 1\% of the training data for downstream training.
Here, SimSiam is especially unstable on MNIST after 100 epochs.
On both data sets, the autoencoder performs better than SimSiam.
Besides, our proposed SidAE model yields better results than the fully-supervised baseline.
The results for using 100\% of the training data for downstream training yield the same outcome and are hence not shown.

\paragraph{Fine-tuning the Downstream Task.}
So far, we kept the pre-trained weights of the encoder frozen when training the downstream classifier.
This is useful to evaluate the capabilities of the feature extractors from the pre-training procedure.
We now fine-tune the weights of the encoder during supervised downstream training.
We set $w = 0.125$ within SidAE due to the previous experiment (Figure~\ref{fig:CIFAR-10_weight_w}).

Figure~\ref{fig:Finetuning} shows the downstream classification accuracies on CIFAR-10 when we allow for fine-tuning, i.e., allowing the backbone to adapt to the downstream task.
As before, our proposed SidAE model outperforms both self-supervised baselines in almost all cases.
In contrast to earlier experiments, all self-supervised models now surpass the supervised baseline.
This shows some form of self-supervised pre-training is beneficial not only when a small amount of labeled data is available, but also if there is a good amount of labeled data.

\begin{figure}[t]
    \centering
    \includegraphics[width=.425\textwidth]{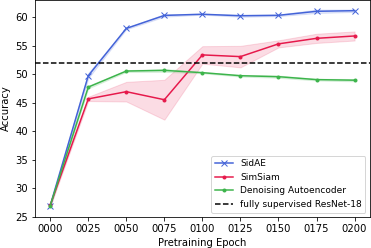}  
    \hfill
    \includegraphics[width=.425\textwidth]{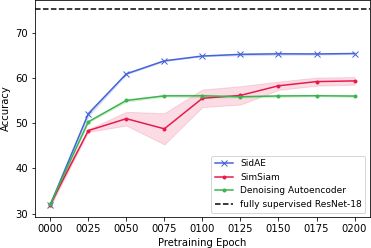}
    \caption{Classification accuracies on STL-10 (left) and CIFAR-10 (right) using an encoder pre-trained with the unlabeled training data of STL-10.
    }
    \label{fig:STL-10}
\end{figure}

\paragraph{Results on STL-10.}
We now pre-train the backbone on the unlabeled training data of STL-10 and use the frozen backbone as a feature extractor for the classification task on STL-10 and also on CIFAR-10.
This allows us to evaluate whether the feature extraction can be pre-trained on data sets with similar characteristics.
We set $w = 0.25$ for SidAE as it yielded the highest downstream accuracy on both data sets.
Similar results are obtained for $w=0.125$.
The classification accuracies on STL-10 and \mbox{CIFAR-10} are depicted in Figure~\ref{fig:STL-10} on the left- and right-hand side, respectively.
For STL-10, we can observe that SidAE and SimSiam both outperform the fully-supervised baseline while the denoising autoencoder falls short.
The situation differs for CIFAR-10, where the fully-supervised baseline is still better.
However, note that the performance of SidAE is very similar to the case when the backbone was pre-trained on CIFAR-10 itself (Figure~\ref{fig:cifar}).
Overall, SidAE performs much better than both self-supervised baselines.

\section{Related Work}

Many models build on Siamese networks \cite{chen2020simple,he2020momentum,grill2020bootstrap,chen2021exploring,liu2022bridging}.
The idea is to make the model learn that two different versions of one object represent the same object. 
The main challenge is to avoid collapsing representations, i.e., all inputs map to the same feature vector.
One widely applied strategy is to use a contrastive loss like InfoNCE \citep{oord2018representation,misra2020self,chen2020simple}, which requires a large amount of negative samples for good performance.
Various approaches have been presented in this context, such as using large batch sizes \citep{chen2020simple}, a memory bank to store negatives \citep{wu2018unsupervised,misra2020self}, or a \emph{queue} of negatives retaining negatives from previous batches \citep{he2020momentum}.
Another approach avoids pairwise comparison by mapping embeddings of images to prototype vectors found by online clustering \cite{caron2020unsupervised}.
Those alleviate but do not solve the issues of dealing with negative samples.
Newer methods follow simpler ways to address the collapsing issue.
BYOL \citep{grill2020bootstrap} circumvents the need for negative samples by applying a momentum encoder, where the weights in one of the branches of the Siamese network are a lagging moving average of the other one.
This makes it difficult for the model to converge to a collapsed state but makes the optimization process more complex.
SimSiam \cite{chen2021exploring} uses a stop-gradient operation on one of the branches instead, showing that it is sufficient to prevent collapsing solutions.
More recently, \cite{liu2022bridging} show that the stop-gradient operation implicitly introduces a constraint that encourages feature decorrelation, explaining why the simple architecture of SimSiam performs well.
\looseness=-1

\section{Conclusion}

We proposed combining the benefits of Siamese networks and denoising autoencoders for learning meaningful data representations.
To achieve this, we introduced a new model called SidAE (\underline{Si}amese \underline{d}enoising \underline{a}uto\underline{e}ncoder).
We empirically evaluated the representations learned by our model in various scenarios on several real-world data sets using classification as a downstream task.
There, we first pre-trained our model on unlabeled training data in a self-supervised fashion before using the extracted features for learning the supervised classifier.
In our experiments, we compared SidAE to SimSiam and a denoising autoencoder.

SidAE consistently outperformed both self-supervised baseline models in terms of mean downstream classification accuracy in all our experiments.
Furthermore, compared to SimSiam, SidAE leads to results that are more stable regarding initialization seeds and the amount of used pre-training epochs.

\subsubsection{Acknowledgements} This work was partially supported by the Wallenberg AI, Autonomous Systems and Software Program (WASP) funded by the Knut and Alice Wallenberg Foundation.

\bibliographystyle{splncs04}
\bibliography{literature}

\end{document}